%% file: main.tex
\documentclass[runningheads]{llncs}

\usepackage[year=2026]{eccv}

\usepackage{eccvabbrv}

\usepackage{graphicx}
\usepackage{booktabs}
\usepackage[accsupp]{axessibility}
\usepackage{xcolor}
\usepackage{amssymb}
\usepackage{tikz}

\usepackage[breaklinks,colorlinks,citecolor=eccvblue]{hyperref}

\newsavebox{\MailIconBox}
\newcommand{\MailIcon}{\usebox{\MailIconBox}}

\begin{document}

\sbox{\MailIconBox}{%
  \tikz[
    baseline=-0.5ex,
    x=1.15ex,
    y=0.78ex,
    line width=0.32pt
  ]{%
    \draw[draw=black,fill=white] (0,0) rectangle (1,1);
    \draw[draw=black] (0,1) -- (0.5,0.42) -- (1,1);
  }%
}


\title{TopoField: Technical Report for ECCV 2026 SceneFun3D Challenge}

\titlerunning{TopoField for SceneFun3D Track 1}

\author{%
  Zhiwei Chen\textsuperscript{*}~~~~
  Zhiheng Fu\textsuperscript{*}~~~~
  Zixu Li\textsuperscript{\textdagger}~~~~
  Mingyu Zhang\\
  Jiale Huang~~~~~
  Yupeng Hu\textsuperscript{\MailIcon}
}

\authorrunning{Chen et al.}

\institute{Shandong University\\[1mm]
  \texttt{\footnotesize\color{black!80} 
\{zivczw,~fuzhiheng8,~lizixu.cs,~zhmingyu.cs\}@gmail.com}
\\
\texttt{\footnotesize\color{black!80}
huangjiale.cs@outlook.com,~huyupeng@sdu.edu.cn
}
  }

\maketitle


\begingroup
  \renewcommand{\thefootnote}{*}
  \footnotetext[0]{These authors contributed equally.}

  \renewcommand{\thefootnote}{\textdagger}
  \footnotetext[0]{Project Leader: Zixu Li.}

  \renewcommand{\thefootnote}{\MailIcon}
  \footnotetext[0]{Corresponding Author: Yupeng Hu.}
\endgroup


\input{Section/0_abstract}
\input{Section/1_intro}
\input{Section/2_related_work}
\input{Section/3_method}
\input{Section/4_experiment}
\input{Section/5_conclusion}


\bibliographystyle{splncs04}
\bibliography{main}

\end{document}

%% file: Section/0_abstract.tex
\begin{abstract}
Functional segmentation requires locating small interactive parts, such as knobs, buttons, and sockets, in indoor point clouds and returning confidence-ranked instance masks. A point-wise semantic predictor produces a continuous probability field, whereas the benchmark evaluates a discrete set of scored instances. This creates a substantial field-to-instance gap. We present \textsc{TopoField}, a deterministic decoder built on top of VOLT. TopoField first converts 13 test-time forward passes into a semantic consensus field, then decodes instance candidates from the consensus geometry, and finally ranks the candidates in a class-aware manner. The augmentation stage retains the processed-point identity of every fragment prediction, accumulates probabilities in a common point array, normalizes by coverage count, and fuses the resulting caches. The decoding stage separates high-confidence cores from broader support regions, constructs a symmetric radius-pruned point graph, and uses a multi-source graph-geodesic partition to split support regions that contain multiple seeds. On 561 ground-truth instances across 30 validation scenes, an IoU-0.5-valid candidate exists for 67.2\% of instances, and top-1 coverage is 20.0\%. These complementary measurements motivate joint attention to candidate construction and ranking. TopoField achieves 28.286\% AP, 41.383\% AP$_{50}$, and 47.558\% AP$_{25}$. TopoField achieved \textbf{second place} on the ECCV 2026 SceneFun3D Challenge.

\keywords{3D point clouds \and functional interaction element segmentation \and instance decoding \and test-time augmentation \and graph-geodesic partitioning}
\end{abstract}

%% file: Section/1_intro.tex
\section{Introduction}
\label{sec:intro}

SceneFun3D~\cite{scenefun3d} asks a model to locate nine categories of functional interaction elements, such as knobs, buttons, handles, and sockets, in complete indoor laser scans. Each prediction requires a category, a point mask, and a confidence score. These elements occupy only a small portion of their host objects, and adjacent elements can share material, color, and local geometry. The task therefore requires not only recognizing an interactable object, but also recovering the spatial extent of the part that affords an action. Related efforts emphasize fine-grained video understanding~\cite{song5}, region-focused visual representations~\cite{OFFSET}, and entity-level alignment in composed retrieval~\cite{TEMA,PAIR,MEDIAN}.

Two coupled difficulties arise after a point-wise semantic predictor such as VOLT~\cite{volt}. Small and partly occluded parts make per-point probabilities noisy: a high threshold truncates weak boundaries, whereas a low threshold can merge the target with its host or a neighbor. In addition, the predictor returns a continuous probability field, while the benchmark evaluates discrete masks sorted by confidence. Even with reasonable semantics, proposal construction and ordering can therefore cause a large field-to-instance error.

We quantify this gap on 30 validation scenes containing 561 evaluable instances. The examined pool has an IoU-0.5-valid candidate for 67.2\% of the instances, and top-1 coverage is 20.0\%. These complementary statistics characterize candidate-pool coverage and the overlap achieved by top-ranked candidates.

We present \textsc{TopoField}, whose method follows the two stages in Fig.~\ref{fig:pipeline}. \textbf{Stage~1, TTA-based Semantic Consensus}, applies the shared predictor to 13 test-time forward passes over 12 distinct rotation--scale outcomes. Each fragment retains a processed-point index that scatters its softmax output into a common point-by-class array; accumulated probabilities are divided by coverage count and the independent caches are renormalized and fused. \textbf{Stage~2, Geometry-aware Instance Decoding}, selects a class-specific operator from argmax connectivity, threshold-support connectivity, or a seeded graph basin. The basin branch separates high-confidence cores from broader supports, builds a symmetric radius-pruned 32-nearest-neighbor graph, and assigns support points to seed components by accumulated graph-path length. Candidates are then scored and suppressed class by class in the original scan point order.

Our contributions are as follows.
\begin{itemize}
  \item We formulate the gap between point-wise semantics and the benchmark's ranked instance-set output, and quantify it through complementary any-candidate and top-$k$ coverage statistics.
  \item We organize the system as two explicit stages that mirror its implementation: index-aligned test-time semantic consensus and geometry-aware instance decoding. The first stage makes point identity and cache fusion explicit; the second exposes core selection, graph construction, geodesic seed assignment, and candidate ranking.
  \item The parameter-free decoder achieves 28.286\% AP, 41.383\% AP$_{50}$, and 47.558\% AP$_{25}$.
\end{itemize}

%% file: Section/2_related_work.tex
\section{Related Work}
\label{sec:related_work}

\textbf{3D instance segmentation.}
PointGroup~\cite{pointgroup} groups point-wise semantics with center offsets in two coordinate spaces, while SoftGroup~\cite{softgroup} reduces hard-label error propagation with soft semantic grouping. Query-based methods such as Mask3D~\cite{mask3d} and SPFormer~\cite{spformer} learn instance or superpoint queries and directly predict masks. These methods jointly learn representation, grouping, and refinement~\cite{HAIS,OneFormer3D}. In visual tracking, complementary approaches explore window-based attention~\cite{song1}, correlative masked modeling~\cite{song2}, and query-gated feature selection~\cite{lv2026gatemot}. TopoField converts a fixed semantic predictor's output into a ranked instance set through deterministic grouping and confidence-based ranking.

\noindent \textbf{Functional interaction element segmentation.}
SceneFun3D~\cite{scenefun3d} extends indoor scene understanding from complete objects to functional parts such as handles, knobs, and buttons. These parts are small, often partly occluded, and geometrically similar to their neighbors~\cite{PartNet,Where2Act}. Related work on composed video retrieval models shared and differential semantics~\cite{REFINE} and calibrates directional anchors in composed representations~\cite{ReTrack}. VOLT~\cite{volt} supplies the point-wise semantics in our system; TopoField constructs and ranks instance proposals from these semantic predictions. In video tracking, semantic consistency strengthens point correspondence~\cite{song4}, while temporally coherent object flow supports multi-frame object association~\cite{song6}.

\noindent \textbf{Graph-based region partitioning.}
Watershed-like methods propagate labels from seeds under a local adjacency structure~\cite{watershed}. A sparse point graph constrains propagation to 3D neighborhoods and prevents unreachable regions from borrowing labels across empty space~\cite{SPG,DGCNN,SPT}. Graph-structured relations also support heterogeneous nearest-neighbor retrieval~\cite{STABLE} and collaborative motion reasoning in multi-object tracking~\cite{song13}. TopoField applies this idea to a class-conditioned support: cores supply seeds, a radius-pruned symmetric kNN graph defines paths, and multi-source shortest paths assign support points to seeds. Connected components first isolate disconnected supports; geodesic propagation then handles connected supports with several cores.

\noindent \textbf{Test-time augmentation.}
Test-time augmentation reduces variance by transforming the input and aggregating the corresponding outputs~\cite{tta,pointnet_tta,qi2017pointnet++,shanmugam2021better}. For per-point prediction, point identity must be preserved across augmentation and test-time voxelization. Stage 1 therefore scatters probabilities through retained point indices, normalizes by coverage, and fuses the caches before Stage 2 consumes the consensus field. Complementary evidence is also integrated through coupled multimodal state-space models~\cite{song3}. Egocentric video understanding systems further explore category-conditioned inference and test-time consistency~\cite{EgoAdapt}, as well as capability-based reasoning routes~\cite{OmniEgo-R2}.

\noindent \textbf{Noise-aware visual matching.}
Reliable visual matching depends on identifying task-relevant evidence and estimating the confidence of candidate associations. Research on noisy correspondence studies these issues in image–text retrieval, where training pairs can contain incorrect semantic associations. Huang et al.~\cite{NCR} introduce the Noisy Correspondence Rectifier, which partitions training pairs according to their estimated reliability and adaptively rectifies their correspondence through co-teaching. Qin et al.~\cite{DECL} develop an evidential learning framework that models correspondence uncertainty and adjusts negative-sample learning through a robust dynamic hinge loss. For composed image retrieval~\cite{ENCODER}, ConeSep~\cite{ConeSep} localizes noisy correspondences through geometric fidelity estimation and performs boundary-based targeted unlearning. Air-Know~\cite{Air-know} integrates expert-prior arbitration with knowledge internalization to guide robust representation learning, while INTENT~\cite{INTENT} combines visual invariant composition with discriminative learning to address cross-modal correspondence noise and modality-inherent interference. Together, these approaches highlight evidence selection and confidence estimation as central components of reliable matching. In TopoField, these considerations motivate the use of semantic consensus, confident core selection, and class-aware candidate scoring for constructing ranked instance predictions.

%% file: Section/3_method.tex
\section{Method}
\label{sec:method}

\begin{figure*}[t]
\centering
\includegraphics[width=\linewidth]{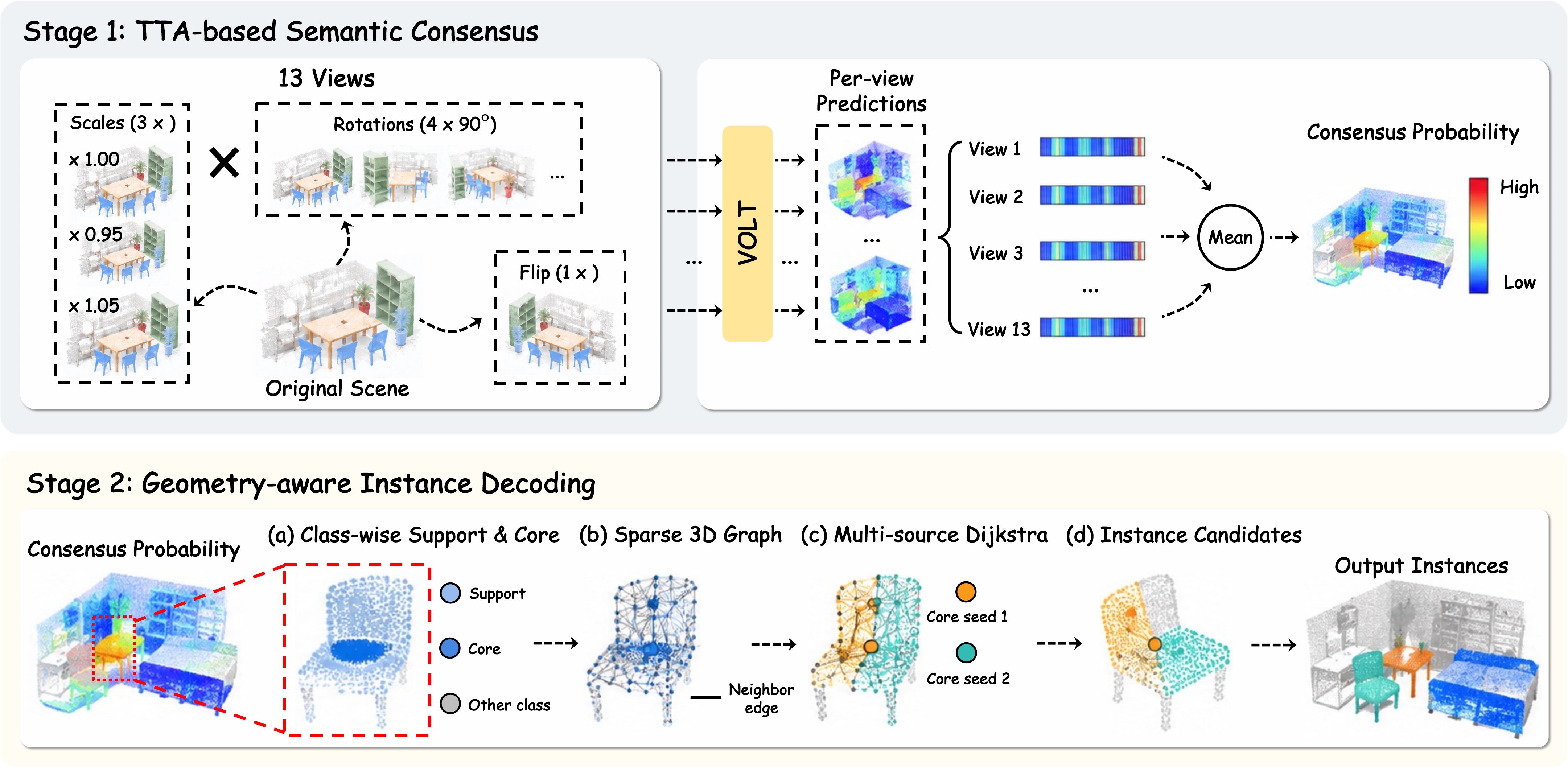}
\caption{TopoField overview. Stage~1 aggregates index-aligned point probabilities from 13 test-time forward passes into a semantic consensus field. Stage~2 applies class-wise initialization, sparse-graph geodesic propagation, and candidate scoring.}
\label{fig:pipeline}
\end{figure*}

\textsc{TopoField} is a deterministic decoder after a point-wise predictor. Stage~1 stabilizes the probability field and resolves index misalignment between augmented inputs; Stage~2 converts the field into instance masks and scores without updating predictor parameters.

\subsection{Problem formulation}
\label{sec:problem_formulation}

Given a scan with $N$ points, the predictor produces $P_i(c)$ for point $i$ and class $c\in\{0,\ldots,9\}$, where class $0$ is background and classes $1,\ldots,9$ are target categories. The decoder returns
\begin{equation}
\label{eq:task_output}
\mathcal{Y}=\{(M_\ell,c_\ell,s_\ell)\}_{\ell=1}^{L},
\qquad
M_\ell\subseteq\{1,\ldots,N\},
\qquad
c_\ell\in\{1,\ldots,9\},
\end{equation}
where $M_\ell$ uses the original point order and $s_\ell$ orders candidates. Only foreground instances are emitted.

\subsection{Stage 1: TTA-based Semantic Consensus}
\label{sec:tta_consensus}

\paragraph{Augmented forward passes.}
The released profile runs 13 forward passes. Its first twelve are the Cartesian product of four vertical-axis rotations, $\rho\in\{0,\frac12,1,\frac32\}$, and three scales, $s\in\{0.95,1.00,1.05\}$. The thirteenth is a double-axis flip. Flipping both horizontal axes equals a 180-degree rotation, so the passes contain 12 distinct spatial transforms.

\paragraph{Index-aligned accumulation.}
For pass $a$ and voxelized fragment point $j$, let $\pi_a(j)$ be its retained processed-point index. Test-time voxelization can place several fragments under the same index. We accumulate softmax probabilities in a common $N\times10$ array
\begin{equation}
\label{eq:scatter_accumulate}
\mathbf{A}^{(a)}_i=\sum_{j:\pi_a(j)=i}\mathbf{p}^{(a)}_j,
\qquad
v^{(a)}_i=\sum_{j:\pi_a(j)=i}1,
\end{equation}
and divide by the number of contributions to each processed point:
\begin{equation}
\label{eq:scatter_normalize}
\tilde{\mathbf{P}}^{(a)}_i=\frac{\mathbf{A}^{(a)}_i}{v^{(a)}_i}.
\end{equation}
Retained processed-point indices align predictions across coordinate and normal transformations, enabling direct accumulation in the common point array.

\paragraph{Cache fusion.}
Each pass writes an independent cache. For positive weights $w_a$ and $W=\sum_a w_a$, we combine the caches and normalize each row:
\begin{equation}
\label{eq:tta_consensus}
\bar P_i(c)=\frac{\sum_{a=1}^{13}(w_a/W)\tilde P^{(a)}_i(c)}
{\sum_{k=0}^{9}\sum_{a=1}^{13}(w_a/W)\tilde P^{(a)}_i(k)}.
\end{equation}
All 13 passes receive equal weights. Since index 12 duplicates index 2 geometrically, their combined weight is $2/13$.

\subsection{Stage 2: Geometry-aware Instance Decoding}
\label{sec:geometry_decoding}

\paragraph{Class-conditioned route.}
The class-conditioned decoder selects among argmax connectivity, foreground-threshold connectivity, and a seeded graph basin for each category. The basin branch in Stage~2 of Fig.~\ref{fig:pipeline} uses confident cores and graph connectivity to partition support regions.

\paragraph{(a) Class-wise support and core.}
For each foreground class $c$, let
\begin{equation}
\label{eq:foreground_argmax}
\mathcal F_c=\left\{i:c=\arg\max_{c'\in\{1,\ldots,9\}}\bar P_i(c')\right\}.
\end{equation}
Given support threshold $\tau_s$ and core threshold $\tau_h$, with $\tau_h\geq\tau_s$, define
\begin{equation}
\label{eq:support_core_regions}
S_c=\{i\in\mathcal F_c:\bar P_i(c)\geq\tau_s\},
\qquad
K_c=\{i\in S_c:\bar P_i(c)\geq\tau_h\}.
\end{equation}
The support retains weak boundary evidence while the core supplies reliable seeds. The representative recipe uses $\tau_s=0.15$, $\tau_h=0.40$, at least five support points, and at least three core points.

\paragraph{(b) Sparse 3D graph.}
We build a graph $G_c=(S_c,E_c)$ only over the support points. Let $\mathcal N_k(i)$ be the $k$-nearest-neighbor set of $i$ in $S_c$. We take the union of outgoing and incoming neighbor relations and remove long edges:
\begin{equation}
\label{eq:graph_edges}
E_c=\{\{i,j\}:j\in\mathcal N_k(i)\ \text{or}\ i\in\mathcal N_k(j),\;\|p_i-p_j\|_2\leq r\}.
\end{equation}
The implementation uses $k=32$ and weight $w_{ij}=\|p_i-p_j\|_2$, bounded below by $10^{-12}$. The resulting sparse, symmetric graph constrains propagation to local 3D neighborhoods.

\paragraph{(c) Multi-source geodesic assignment.}
Connected components of $G_c$ are formed first. Inside each support component, the core points induce seed components $Q_{cjm}$. If a support component contains more than one seed component, we run the multi-source Dijkstra step from all seeds and assign each support point to the seed with the smallest accumulated graph distance:
\begin{equation}
\label{eq:seed_basin}
a_{cj}(i)=\arg\min_m\min_{q\in Q_{cjm}}d_{G_c[C_{cj}]}(i,q).
\end{equation}
The resulting candidate is
\begin{equation}
\label{eq:instance_candidates}
M_{cjm}=\{i\in C_{cj}:a_{cj}(i)=m\}.
\end{equation}
With one seed component, the whole support component forms a single candidate.

\paragraph{(d) Candidate ranking and suppression.}
For a materialized candidate $M$, the primary score is the mean consensus probability of its class:
\begin{equation}
\label{eq:candidate_score}
s(M,c)=\frac{1}{|M|}\sum_{i\in M}\bar P_i(c).
\end{equation}
The decoder also supports quantile and hybrid pooling, and a class-wise size factor,
\begin{equation}
\label{eq:size_score}
s'(M,c)=s(M,c)\left(\frac{|M|}{r_c}\right)^{\alpha_c},
\end{equation}
where $r_c$ is a reference point count and $\alpha_c=0$ recovers the unscaled score. Exact duplicate masks keep the highest-scoring copy. For overlapping same-class candidates, soft NMS uses
\begin{equation}
\label{eq:soft_nms}
s_b\leftarrow s_b\exp\left(-\frac{\operatorname{IoU}(M_a,M_b)^2}{\sigma}\right).
\end{equation}
Candidates are filtered and suppressed class by class, and the final scene list is returned in original point order.

\subsection{Implementation details}
\label{sec:implementation}

VOLT predicts at a 1\,cm voxel resolution from XYZ, RGB, and normals. It has 12 transformer blocks, 768-dimensional embeddings, and 12 heads, initialized from ScanNet200 weights. Training uses AdamW ($3\times10^{-4}$, weight decay 0.05), label-smoothed cross-entropy, multiclass Lov\'asz loss, and EMA (0.999). Test inference uses frozen probability caches without gradient updates. Unless stated otherwise, Stage~1 is fixed while Stage~2 is varied.

%% file: Section/4_experiment.tex
\section{Experiments}
\label{sec:experiments}

\subsection{Setup}

SceneFun3D~\cite{scenefun3d} contains 160 training, 30 validation, and 30 test scenes over nine categories. We train VOLT with XYZ, RGB, and normals at 1\,cm. Stage~1 is used only at inference. The official metric is AP over IoU 0.50--0.95, with AP$_{50}$ and AP$_{25}$ also reported. $R_{\mathrm{any}}$ is the fraction of ground-truth instances whose candidate pool contains an IoU-0.5-valid mask; $R_k$ restricts this test to the class-wise top-$k$ candidates.

\subsection{Candidate diagnosis}

Using fixed epoch-80 EMA 13-TTA caches, Table~\ref{tab:candidate_diagnosis} shows a 47.2-point gap between any-candidate and top-1 coverage. The two measurements distinguish candidate-pool coverage from top-ranked mask quality and provide complementary criteria for evaluating Stage~2.

\begin{center}
\small
\captionof{table}{Candidate diagnosis on 30 validation scenes (561 ground-truth instances). Values are percentages.}
\label{tab:candidate_diagnosis}
\begin{tabular}{lc}
\toprule
Diagnostic & Value \\
\midrule
Foreground-point precision / recall & 68.4 / 60.1 \\
$R_{\mathrm{any}}$@0.5 & 67.2 \\
$R_1$@0.5 & 20.0 \\
$R_3$@0.5 & 38.3 \\
$R_5$@0.5 & 46.3 \\
\bottomrule
\end{tabular}
\end{center}

\subsection{Validation and test results}

Table~\ref{tab:validation_configs} reports validation results for successive decoder configurations. Support/core selection gives the largest improvement, and graph geodesic partitioning further separates adjacent candidates. Class-conditioned decoding raises AP from 37.60 to 38.45 on the development split and from 41.86 to 42.31 on the holdout split, reaching 39.67 AP across all 30 validation scenes.

\begin{center}
\small
\captionof{table}{Validation configuration search (epoch-80, 13-TTA). AP values are percentages.}
\label{tab:validation_configs}
\begin{tabular}{lccc}
\toprule
Configuration & Development (20) & Holdout (10) & Full validation (30) \\
\midrule
Baseline & 35.21 & 39.47 & 36.58 \\
+ Dual-threshold & 36.89 & 41.02 & 38.14 \\
+ Graph geodesic & 37.60 & 41.86 & 38.92 \\
+ Class-conditioned & 38.45 & 42.31 & 39.67 \\
\bottomrule
\end{tabular}
\end{center}

Table~\ref{tab:test_results} reports official submissions from two EMA checkpoints with robust and aggressive post-processing. Each checkpoint uses its associated saved seed configuration. The epoch-100 aggressive configuration achieves 28.286\% AP, 41.383\% AP$_{50}$, and 47.558\% AP$_{25}$.

\begin{center}
\small
\captionof{table}{Official-test submission results. AP, AP$_{50}$, and AP$_{25}$ are percentages.}
\label{tab:test_results}
\begin{tabular}{lcccc}
\toprule
EMA epoch & Post-processing & AP & AP$_{50}$ & AP$_{25}$ \\
\midrule
80 & Robust & 27.919 & 41.373 & 45.878 \\
80 & Aggressive & 27.640 & 41.103 & 46.364 \\
100 & Robust & 28.270 & 40.847 & 46.761 \\
100 & Aggressive & \textbf{28.286} & \textbf{41.383} & \textbf{47.558} \\
\bottomrule
\end{tabular}
\end{center}

\subsection{Discussion}

\textbf{Decoder components.} The successive configurations in Table~\ref{tab:validation_configs} increase development-split AP by 1.68 points with support/core selection, 0.71 points with graph geodesic partitioning, and 0.85 points with class-conditioned decoding. Together, these components raise full-validation AP from 36.58 to 39.67.

\textbf{Method behavior.} The two stages perform complementary operations. Stage~1 aggregates index-aligned point predictions into a semantic consensus field. Stage~2 constructs and ranks instance masks using class-conditioned connectivity and graph-geodesic partitioning. Support/core selection retains boundary evidence and supplies confident seeds, while geodesic assignment partitions multi-core support regions according to their 3D connectivity.

%% file: Section/5_conclusion.tex
\section{Conclusion}
\label{sec:conclusion}

We presented TopoField, a two-stage deterministic decoder for functional interaction element segmentation. Stage~1 turns multiple test-time forward passes into a stable semantic field by retaining processed-point indices, accumulating softmax probabilities in a common point array, normalizing repeated fragment contributions by coverage, and fusing independent caches. Stage~2 uses class-conditioned instance decoding; its seeded graph-basin branch separates a high-confidence core from a broader support region, constrains propagation with a symmetric radius-pruned point graph, and assigns support points to seed components by graph-geodesic distance. The validation analysis characterizes candidate coverage and ranking, and the combined decoder achieves 39.67\% validation AP and 28.286\% official-test AP. The decoder adds no learnable parameters to the point-wise predictor.